# Embedding Words as Distributions with a Bayesian Skip-gram Model

**Arthur Bražinskas**[1]     **Serhii Havrylov**[2]     **Ivan Titov**[1,2]
[1]ILLC, University of Amsterdam
[2]ILCC, School of Informatics, University of Edinburgh
arthur.brazinskas@gmail.com  s.havrylov@ed.ac.uk  ititov@inf.ed.ac.uk

## Abstract

We introduce a method for embedding words as probability densities in a low-dimensional space. Rather than assuming that a word embedding is fixed across the entire text collection, as in standard word embedding methods, in our Bayesian model we generate it from a word-specific prior density for each occurrence of a given word. Intuitively, for each word, the prior density encodes the distribution of its potential 'meanings'. These prior densities are conceptually similar to Gaussian embeddings of Vilnis and McCallum (2015). Interestingly, unlike the Gaussian embeddings, we can also obtain context-specific densities: they encode uncertainty about the sense of a word given its context and correspond to the approximate posterior distributions within our model. The context-dependent densities have many potential applications: for example, we show that they can be directly used in the lexical substitution task. We describe an effective estimation method based on the variational autoencoding framework and demonstrate the effectiveness of our embedding technique on a range of standard benchmarks.

## 1 Introduction

Distributed representations of words induced from large unlabeled text collections have had a large impact on many natural language processing (NLP) applications, providing an effective and simple way of dealing with data sparsity. Word embedding methods typically represent words as vectors in a low-dimensional space (Deerwester et al., 1990; Collobert et al., 2011; Mikolov et al., 2013; Pennington et al., 2014). In contrast, we encode them as probability densities (see Figure 1). Intuitively, the densities represent the distributions over possible meanings[1] of a word. Representing a word as a distribution provides many potential benefits. For example, such embeddings let us encode generality of terms (e.g., kakapo is a hypernym of bird), characterize uncertainty about semantic properties of the corresponding referent (e.g., a proper noun, such as John, encodes little about the person it refers to) or represent polysemy (e.g., kiwi may refer to a fruit, a bird or a New Zealander).

The main inspiration for this work is Gaussian embeddings (*word2gauss*, W2G) introduced by Vilnis and McCallum (2015). They represent words as Gaussian distributions and directly optimize an objective expressed in terms of divergences (e.g., the Kullback-Liebler divergence) between the distributions. In contrast, we approach the problem from the generative Bayesian perspective. Though, as in W2G, context-agnostic densities are present in our model (they correspond to data-dependent priors), unlike W2G, we can also perform posterior inference and obtain context-specific densities. These posterior densities encode semantic properties of a word in a given context. For example, in Figure 1, when 'kiwi' appears in a context suggesting the 'bird' sense, the posterior (represented by the shaded ellipsoid) becomes more 'peaky' and moves towards the representation of word 'bird'. We use a lexical substitution task (McCarthy and Navigli, 2007) to demonstrate that the posterior densities are effective in predicting potential replacements of a word given a context. Importantly, though the Gaussian assumption for context-agnostic embeddings is questionable (e.g., polysemous words would need multimodal distributions to accurately represent their meanings), the same assumption for the context-specific posterior densities is more reasonable: the word is likely to be disambiguated by the provided context and hence does not require complex families of distributions.

---

[1]We use the term 'meaning' somewhat liberally in this work: we are not arguing that embeddings capture actual meanings.

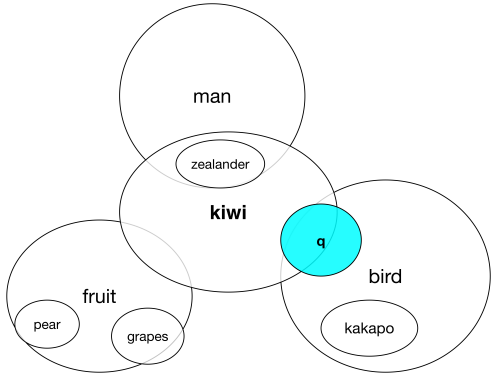

Figure 1: An idealized illustration of density embeddings. Unshaded ellipsoids encode prior densities. The shaded ellipsoid corresponds to the posterior for 'kiwi' when it appears in a context indicating that 'kiwi' refers to a bird.

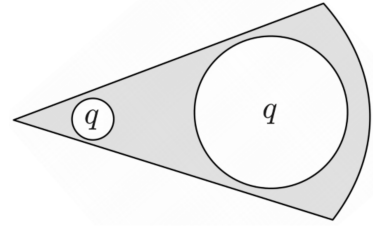

Figure 2: Shaded cone is a fixed angle, and ellipses are approximate posterior Gaussian distributions. The corner of the cone is at the origin.

In principle, using densities to represent words provides a natural way of encoding entailment: the decision regarding entailment relation can be made by testing the level sets of the distributions for soft inclusion. For example, in Figure 1, the ellipse for 'kakapo' lies within the ellipse for 'bird'. Vilnis and McCallum (2015) proposed to use the KL divergence to detect entailment. In our analysis, we observe that, though the covariances indeed encode information relevant to entailment, their direct use is somewhat problematic both with W2G and with our model.

We are not the first to propose a Bayesian version of word embedding methods (Zhang et al., 2014; Sakaya, 2015; Barkan, 2017). However, our approach is crucially different from the previous work. The previous methods can be regarded as applications of Bayesian matrix factorization (BMF) (Salakhutdinov and Mnih, 2008) to word co-occurrence matrices. Hence they assume that every word is associated with a fixed (but unknown) real-valued vector which is shared across the entire text collection. Instead, in our approach, we acknowledge that word meaning inherently depends on a context and do not assume that any fixed vector exists at type level: we draw it at token level (i.e. for each word occurrence) from a parameterized word-specific prior distribution. Consequently, unlike us and also unlike densities in W2G, the posteriors in previous Bayesian embeddings models encode uncertainty about the embedding parameters; they will converge to a delta distribution as the amount of data increases. In contrast, our densities encode the distribution of senses and, as confirmed in our experiments, do not follow this trend.

More formally, our model can be regarded as a form of coupled matrix factorization where each sliding window is factorized individually, but priors for words are shared across all the windows. We describe an effective estimation method based on the variational autoencoding (VAE) framework (Kingma and Welling, 2014). The context-specific densities are provided by the encoder component of VAE (i.e. the inference network). Our main contributions can be summarized as follows:[2]

- we proposed a Bayesian model for embedding words as probability densities;
- we derived a computationally-efficient inference algorithm, which, as a by-product, yields context-sensitives densities;
- we demonstrate their effectiveness, including on the lexical substitution task.

## 2 Bayesian Skip-gram

In this section, we provide detailed description of the novel model. To motivate our Bayesian extension of the Skip-gram model, consider the polysemous word 'kiwi' (Figure 1). Its meaning changes depending on the context: for example, when it appears in 'I like apples, kiwi, and bananas', we can deduce that 'kiwi' refers to a fruit. Polysemy cannot be effectively captured with a single representation induced by standard word embedding methods (e.g., skipgram (Mikolov et al., 2013)) or even with a single distribution, as in W2G. We also do not want to assume that there is a finite set of discrete senses, as made in multi-sense mixture models (Li and Jurafsky, 2015; Neelakantan et al., 2014). Sense distinctions are often gradual and discrete senses are only clusters that approximate underlying meaning distributions (Kilgarriff, 1997; Erk and McCarthy, 2009). Our Bayesian Skip-gram (BSG) model addresses this problem. It predicts a distribution of 'meanings' given a context. Intuitively, if the context is very discriminative, this density becomes very peaky, assigning the entire probability mass to the predicted 'meaning' (in the limit, encoding the word as a point vector). If the context is less informative, the distribution will remain flat, representing uncertainty about the word sense.

---
[2] The implementations of our model is available at https://github.com/ixlan/BSG.

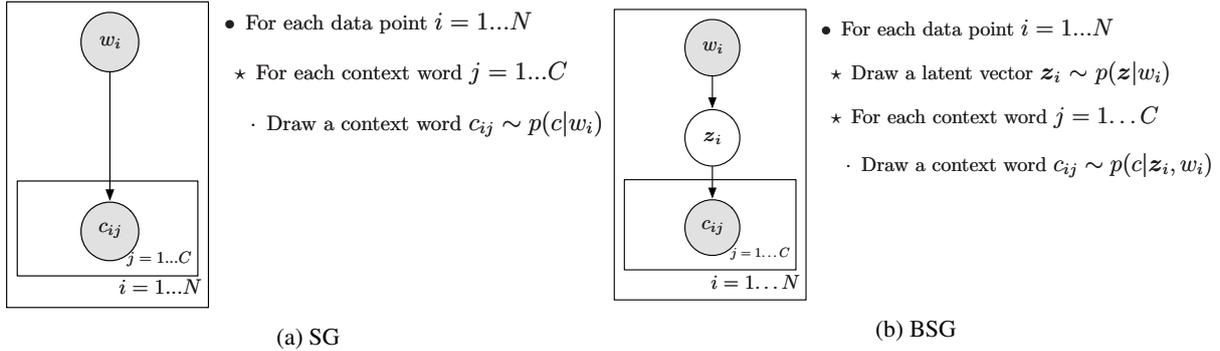

Figure 3: Bayesian networks corresponding to Skip-gram and Bayesian skip-gram.

## 2.1 Generative model

The skip-gram (SG) model aims at maximizing the probability $p_{\boldsymbol{\theta}}(\mathbf{c}|w)$ of words in a context window given its central word. The log probability of each context word is assumed to be proportional to the dot product of its representation and that of the central word (see the graphical model in Figure 3a). In contrast, BSG assumes that the choice of context words is dependent on the context-specific (latent) meaning of the central word (see Figure 3b).

The generative story of BSG is the following: take a word from the dataset (e.g. 'kiwi'), sample its latent meaning $\mathbf{z} \sim p_{\boldsymbol{\theta}}(\mathbf{z}|w)$ (e.g., a vector that encodes the meaning 'bird'), and finally draw context words $c \sim p_{\boldsymbol{\theta}}(c|\mathbf{z})$ (e.g., 'flightless', 'forest', and 'feather').

## 2.2 Model definition

Ideally, we would like to maximize the log-likelihood, which is the sum of the following terms, one per each window:

$$\log p_{\boldsymbol{\theta}}(\mathbf{c}|w) = \log \int \prod_{j=1}^{C} p_{\boldsymbol{\theta}}(c_j|\mathbf{z}) p_{\boldsymbol{\theta}}(\mathbf{z}|w) \mathrm{d}\mathbf{z} \quad (1)$$

where $C$ is the size of the context window $\mathbf{c}$, $c_j$ is a context word for the central word $w$, and $\boldsymbol{\theta}$ are model parameters. Unfortunately, as $p_{\boldsymbol{\theta}}(c_j|\mathbf{z})$ is a neural network, integration over the latent space is intractable. Hence, the marginal log-likelihood and its derivatives cannot be efficiently computed. In our case posterior distribution $p_{\boldsymbol{\theta}}(\mathbf{z}|\mathbf{c}) = p_{\boldsymbol{\theta}}(\mathbf{c}|\mathbf{z})p_{\boldsymbol{\theta}}(\mathbf{z})/p_{\boldsymbol{\theta}}(\mathbf{c})$ is intractable as well. This means that the expectation-maximization (EM) algorithm cannot be used. Instead, we rely on variational inference, specifically the variational auto-encoding (VAE) framework (Kingma and Welling, 2014).

## 2.3 Bayesian Skip-gram ELBO

For BSG model we optimize the variational lower bound of the marginal likelihood:

$$\log p_{\boldsymbol{\theta}}(\mathbf{c}|w) \geq \sum_{j=1}^{C} \mathbb{E}_{q_{\boldsymbol{\phi}}(\mathbf{z}|\mathbf{c},w)}\left[\log p_{\boldsymbol{\theta}}(c_j|\mathbf{z})\right] - \mathbb{D}_{KL}\left[q_{\boldsymbol{\phi}}(\mathbf{z}|\mathbf{c},w) \| p_{\boldsymbol{\theta}}(\mathbf{z}|w)\right] \quad (2)$$

where $q_\phi(\mathbf{z}|\mathbf{c}, w)$ is the approximate posterior distribution, which can be used to infer the representation ('meaning') of the central word $w$ in the context $\mathbf{c}$. For now, we will assume that the approximate posterior is a Gaussian distribution with a diagonal covariance matrix, namely $q_\phi(\mathbf{z}|\mathbf{c}, w) = \mathcal{N}(\mathbf{z}|\boldsymbol{\mu}_q, \boldsymbol{\Sigma}_q)$. The covariance matrix of the approximate posterior encodes uncertainty or generality of meaning of the central word $w$ in the context $\mathbf{c}$. Intuitively, if the meaning is concrete, we would like to get small variance values and large, otherwise.

Another important component of our model is $p_{\boldsymbol{\theta}}(\mathbf{z}|w)$, to which we will refer as a *prior*, and we will assume that it is also a Gaussian distribution with a diagonal covariance[3] and individual parameters for each central word $p_{\boldsymbol{\theta}}(\mathbf{z}|w) = \mathcal{N}(\mathbf{z}|\boldsymbol{\mu}_w, \boldsymbol{\Sigma}_w)$. Fortunately, the KL divergence term between two normal distributions can be expressed in closed form.

## 2.4 Reconstruction error

The most troublesome component of the lower bound is the expected reconstruction term $\mathbb{E}_{q_\phi(\mathbf{z}|\mathbf{c},w)}[\log p_{\boldsymbol{\theta}}(c_j|\mathbf{z})]$ which can be decomposed as shown in Eq. (3):

$$\mathbb{E}_{q_\phi(\mathbf{z}|\mathbf{c},w)}[\log p_{\boldsymbol{\theta}}(c_j|\mathbf{z})] = \mathbb{E}_{q_\phi(\mathbf{z}|\mathbf{c},w)}\left[\log \frac{\exp(f_{\boldsymbol{\theta}}(\mathbf{z}, c_j))}{\sum_{k=1}^{|V|} \exp(f_{\boldsymbol{\theta}}(\mathbf{z}, c_k))}\right] \quad (3)$$

where $p_{\boldsymbol{\theta}}(c_j|\mathbf{z})$ is represented by a neural network with the softmax activation function and $f_{\boldsymbol{\theta}}(\mathbf{z}, c_j)$ is a function that models relationship between latent vector $\mathbf{z}$ and context word $c_j$. In the original skip-gram model it was the dot product. The dot product is based on the angle between two vectors scaled by their norms, which makes a lot of sense with point estimates. However, it is problematic when we use normal distributions. Consider the fixed angle cone and two densities in Figure 2. Both densities encode the same uncertainty about the angle, whereas their variances are very different. Hence, the uncertainty in the dot product can be increased either by moving the density towards the origin or by increasing the variance. In other words, the model has too many degrees of freedom. Consequently, words that are more general will not necessarily have a large variance. We observed this behaviour in our toy experiment.

To address this problem, we propose to use a different form of the function $f$: $f_{\boldsymbol{\theta}}(\mathbf{z}, c_j) = \log\left(\mathcal{N}(\mathbf{z}; \boldsymbol{\mu}_{c_j}, \boldsymbol{\Sigma}_{c_j}) \times p(c_j)\right)$:[4]

$$\mathbb{E}_{q_\phi}\left[\log \frac{\exp(f_{\boldsymbol{\theta}}(\mathbf{z}, c_j))}{\sum_{k=1}^{|V|} \exp(f_{\boldsymbol{\theta}}(\mathbf{z}, c_k))}\right] = \mathbb{E}_{q_\phi}\left[\log\left(\mathcal{N}(\mathbf{z}; \boldsymbol{\mu}_{c_j}, \boldsymbol{\Sigma}_{c_j})p(c_j)\right) - \log \sum_{k=1}^{|V|} \mathcal{N}(\mathbf{z}; \boldsymbol{\mu}_{c_k}, \boldsymbol{\Sigma}_{c_k})p(c_k)\right] \quad (4)$$

Intuitively, scaling with $p(c_j)$ (e.g., the empirical unigram probability) encodes that frequent words are more likely to appear in any context. The Gaussian density evaluates how well a word fits context given the meaning $\mathbf{z}$ of the central word and is more sensitive to the variance, hence addressing the above-mentioned problem.

## 2.5 Encoder

An important component of our model is the inference network $q_\phi(\mathbf{z}|\mathbf{c}, w)$ (also known as an encoder), which approximates the posterior distribution $p_{\boldsymbol{\theta}}(\mathbf{z}|\mathbf{c}, w)$. As we discussed, we make the Gaussian assumption for $q$. Similarly to Kingma and Welling (2014), we use a one-layer feed-forward neural network to compute variance and mean parameters of $q$. Its architecture is shown in Figure 4. The network takes embeddings of the central word and those of context words, passes each of them through a non-linearity (ReLU) and sums the vectors together: $\mathbf{h} = \sum_{j=1}^{C} relu\left(\mathbf{M} \begin{bmatrix} \mathbf{R}_{c_j} \\ \mathbf{R}_w \end{bmatrix}\right)$. The resulting vector is then used to generate $\boldsymbol{\mu}_q$ and $\log \sigma^2$ with a linear model: $\log \sigma_q^2 = \mathbf{W}\mathbf{h} + b_2$; $\boldsymbol{\mu}_q = \mathbf{U}\mathbf{h} + \mathbf{b}_1$. Using logarithm is necessary to ensure that the matrix is positive definite. The encoder can be understood as a parametrized function that produces distributions corresponding to meanings of central words in different contexts. Using this function, we can model infinitely many meanings of words without explicitly storing their representations.

---

[3] Though this assumption is questionable for polysemous words, what is crucial is that in BSG (unlike W2G) we have access to approximate context-specific posterior densities encoding word senses. Nevertheless, using more expressive priors may be beneficial. Expressive data-dependent priors may also lead to over-fitting as using simple priors can be regarded as a form of regularization. We leave investigation of richer priors for future work.

[4] For brevity, we use $q_\phi$ to denote $q_\phi(\mathbf{z}|\mathbf{c}, w)$.

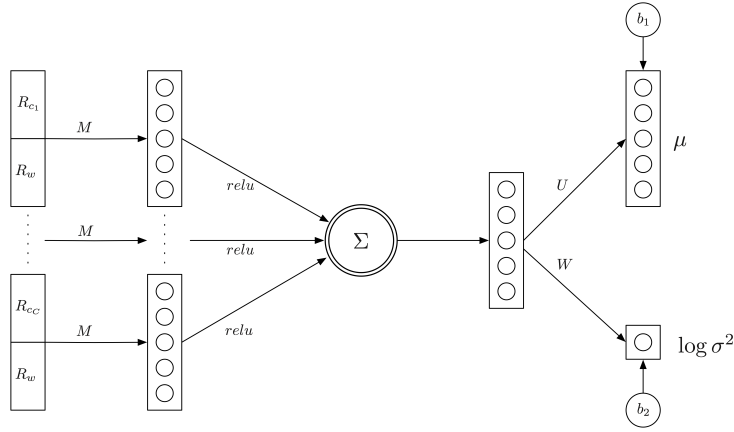

Figure 4: Encoder architecture.

## 2.6 Approximation of the log-partition function

While we can obtain a low-variance estimate of the first part of the expectation from the Eq. (4) using the re-parameterization trick (Kingma and Welling, 2014), the expected log-partition function (the second part) is very computationally demanding, as it involves summation over all words in the vocabulary. In order to scale it to large vocabularies, we compute the lower bound of the log-partition function and use a Monte Carlo (MC) approximation to obtain an unbiased estimate of the expectation over all words:

$$\mathbb{E}_{q_\phi}\left[\log \mathbb{E}_{p(\tilde{c})}\left[\mathcal{N}(\mathbf{z}; \boldsymbol{\mu}_{\tilde{c}}, \boldsymbol{\Sigma}_{\tilde{c}})\right]\right] \geq \mathbb{E}_{q_\phi}\left[\mathbb{E}_{p(\tilde{c})}\left[\log \mathcal{N}(\mathbf{z}; \boldsymbol{\mu}_{\tilde{c}}, \boldsymbol{\Sigma}_{\tilde{c}})\right]\right] \approx \mathbb{E}_{q_\phi}\left[\log \mathcal{N}(\mathbf{z}; \boldsymbol{\mu}_{\tilde{c}_j}, \boldsymbol{\Sigma}_{\tilde{c}_j})\right] \quad (5)$$

The estimate involves only one word $\tilde{c}_j$ ('negative word') which can be easily sampled from $p(\tilde{c})$. By replacing the log-partition function from Eq. (4) with the unbiased estimate of its lower bound derived in Eq. (5), we can get the approximate lower-bound of marginal log-likelihood[5] shown in Eq. (6). The sum $\sum_{(j,k)}$ in the equation is now over pairs of positive $c_j$ and negative $\tilde{c}_k$ context words. Furthermore, we can transform the difference of negative cross-entropies into the difference of Kullback-Leibler divergences by adding and subtracting entropy of $q_\phi(\mathbf{z}|\mathbf{c}, w)$:

$$\begin{aligned}\hat{\mathcal{L}} = \sum_{(j,k)} \left(\mathbb{E}_{q_\phi}\left[\log \mathcal{N}(\mathbf{z}; \boldsymbol{\mu}_{c_j}, \boldsymbol{\Sigma}_{c_j})\right] - \mathbb{E}_{q_\phi}\left[\log \mathcal{N}(\mathbf{z}; \boldsymbol{\mu}_{\tilde{c}_k}, \boldsymbol{\Sigma}_{\tilde{c}_k})\right]\right) - \mathbb{D}_{KL}\left[q_\phi \| p_\theta(\mathbf{z}|w)\right] = \\ \sum_{(j,k)} \left(\mathbb{D}_{KL}\left[q_\phi \| \mathcal{N}(\mathbf{z}; \boldsymbol{\mu}_{\tilde{c}_k}, \boldsymbol{\Sigma}_{\tilde{c}_k})\right] - \mathbb{D}_{KL}\left[q_\phi \| \mathcal{N}(\mathbf{z}; \boldsymbol{\mu}_{c_j}, \boldsymbol{\Sigma}_{c_j})\right]\right) - \mathbb{D}_{KL}\left[q_\phi \| p_\theta(\mathbf{z}|w)\right]\end{aligned} \quad (6)$$

The first term in the above equation suggests maximization of the margin between divergences that involve negative and positive words. We transform it into the hard-margin form (i.e. use the hinge loss, as done in Weston et al. (2013)). This is the final objective function that we maximize:

$$\sum_{(j,k)} \max\left(0, \mathbb{D}_{KL}\left[q_\phi \| \mathcal{N}(\mathbf{z}; \boldsymbol{\mu}_{c_j}, \boldsymbol{\Sigma}_{c_j})\right] - \mathbb{D}_{KL}\left[q_\phi \| \mathcal{N}(\mathbf{z}; \boldsymbol{\mu}_{\tilde{c}_k}, \boldsymbol{\Sigma}_{\tilde{c}_k})\right] + m\right) + \mathbb{D}_{KL}\left[q_\phi \| p_\theta(\mathbf{z}|w)\right] \quad (7)$$

This transformation of the objective is necessary because the KL terms are unconstrained, and, during optimization, the model tends to assign extreme values for the frequent negative words and this has a detrimental effect on the overall performance. The hinge loss solves this problem by setting the gain to zero when the KL term involving a negative context word is larger by a margin than the KL term involving a positive one.

---
[5] Notice that $p(c_j)$ is omitted because it factors out as a constant and does not change the optimum.

| Datasets | BSG | WG(S) | WG(D) | SG |
|---|---|---|---|---|
| MC-30 | 0.71 | 0.69 | 0.70 | **0.72** |
| MEN-TR-3k | **0.73** | 0.72 | 0.71 | 0.72 |
| MTurk-287 | 0.70 | 0.70 | 0.69 | **0.70** |
| MTurk-771 | **0.67** | 0.65 | 0.64 | 0.65 |
| RG-65 | 0.70 | 0.69 | 0.71 | **0.72** |
| RW-STNFRD | 0.43 | 0.43 | 0.42 | **0.44** |
| SIMLEX-999 | **0.35** | 0.34 | 0.34 | 0.34 |
| VERB-143 | 0.32 | **0.38** | 0.29 | 0.36 |
| WS-353-ALL | **0.72** | 0.68 | 0.67 | 0.69 |
| WS-353-REL | **0.68** | 0.66 | 0.65 | 0.65 |
| WS-353-SIM | **0.75** | 0.70 | 0.68 | 0.71 |
| YP-130 | **0.50** | 0.46 | 0.46 | 0.45 |
| Sum | **7.26** | 7.10 | 6.95 | 7.15 |

Table 1: Word similarity benchmarks.

| Model | GAP |
|---|---|
| BSG (Encoder) | **0.461** |
| BSG (Add) | 0.437 |
| BSG (Mult) | 0.439 |
| W2G(S) (Add) | 0.431 |
| W2G(S) (Mult) | 0.432 |
| W2G(D) (Add) | 0.427 |
| W2G(D) (Mult) | 0.427 |
| SG (Add) | 0.426 |
| SG (Mult) | 0.428 |

Table 2: Lexical substitution task. Average precision for SemEval-07 dataset.

The intuition behind the objective is the following: once we generated the posterior distribution $q_\phi(\mathbf{z}|\mathbf{c}, w)$, we optimize parameters to make $\mathbb{D}_{KL}\left[q_\phi(\mathbf{z}|\mathbf{c}, w) \| p_\theta(\mathbf{z}|w)\right]$ small, which can be intuitively understood as a regularization preventing $q_\phi(\mathbf{z}|\mathbf{c}, w)$ ellipsoid from diverging from $p_\theta(\mathbf{z}|w)$ ellipsoid. In addition, we discriminate positive context words from negative ones by comparing their divergences from $q_\phi(\mathbf{z}|\mathbf{c}, w)$.

The expected value of the log-partition function in Eq. (3) is with the negative sign, and because we approximate its lower-bound, the resulting objective function in Eq. (6) is not a lower bound of the likelihood anymore. We still found that proposed approximation works quite well in practice. As an alternative, we considered a Monte Carlo approximation of the partition function proposed by Botev et al. (2017). However, this led to inferior performance on our benchmarks.

## 3 Experiments

In this section we empirically evaluate our approach. First, in subsections 3.2 - 3.4, we use standard benchmarks to evaluate and better understand properties of our context-agnostic embeddings (i.e. the prior distributions $\mathcal{N}(\mathbf{z}; \boldsymbol{\mu}_{c_j}, \boldsymbol{\Sigma}_{c_j})$). In these experiments, we can directly compare BSG to W2G and the vanilla version of skip-gram. In subsection 3.5, we also demonstrate that the context-sensitive posterior distributions $q_\phi(\mathbf{z}|\mathbf{c}, w)$, estimated by the VAE inference network, can be used to select potential substitutes of a word in a given context.

### 3.1 Experimental settings

We train all our models on a concatenation of ukWaC and WaCkypedia (Baroni et al., 2009) corpora, resulting in approximately 3 billion tokens. The embedding dimensionality was set to 100 for all the models. For BSG, we used spherical covariances both for the posterior and prior densities, as they resulted in better performance in preliminary experiments: this is consistent with results reported for W2G in Vilnis and McCallum (2015). In order to enable fair comparison, we used our own implementation of W2G and SG, which shared the hyperparameters with BSG.[6] Throughout the text we use WG(S) and WG(D) to denote W2G with the spherical covariances and W2G with the diagonal covariance matrix, respectively. For additional details see supplementary materials.

---
[6]The original implementation of W2G is not publicly available. Our re-implementation yields stronger results on similarity benchmarks but weaker on the entailment dataset of Baroni et al. (2012). Our implementation is also stronger across the board (including entailment) than the third-party implementation used in Vulić et al. (2017) (https://github.com/seomoz/word2gauss). Athiwaratkun and Wilson (2017) also report W2G numbers very similar to ours.

|       | BBDS |      | BLESS |      |
| ----- | ---- | ---- | ----- | ---- |
| Model | KL   | Cos  | KL    | Cos  |
| BSG   | 76.2 | **75.9** | 20.0 | 20.8 |
| W2G(S)| **77.0** | 75.7 | 18.9 | 20.4 |
| W2G(D)| 76.5 | 74.9 | 18.7 | 20.3 |
| SG    | -    | 75.7 | -     | 20.3 |

Table 3: F1 scores on entailment recognition.

| Model | BBDS | BLESS |
| ----- | ---- | ----- |
| BSG   | 78.23 | **67.34** |
| W2G(S)| 78.41 | 57.50 |
| W2S(D)| 78.05 | 54.58 |
| Freq. Baseline | **78.84** | 55.26 |

Table 4: Entailment directionality detection.

## 3.2 Word similarity

Table 1 presents similarity results computed using the online tool of Faruqui and Dyer (2014). In these experiments, we used only the mean vectors (from the prior) and ignored the covariance information, both for BSG and W2Gs. First, we observe that BSG has a slight edge over both the original SG and also over W2G versions. Second, contrary to observations in Vilnis and McCallum (2015), W2G does not reach a performance of SG in our experiments. As their original implementation is not available, it is hard to pinpoint the reason for the discrepancy. Note that their SG baseline is considerably weaker than ours, so it may be easier to beat. Our SG baselines outperforms their reported results on all similarity datasets (they considered only 8 out of 12 in Table 1) and also in average (0.6248 vs. 0.5724).[7] The results suggest that prior means induced by BSG are indeed effective in capturing semantic properties of words.

## 3.3 Entailment recognition

In this section, we consider the lexical entailment task (i.e. essentially hyponymy detection). Note that, as with word similarity, word context is not provided, and, thus, we cannot showcase the main advantage of our approach: its disambiguation capabilities.

Given an ordered pair of words, the task is to predict if the first word ($w_1$) entails the second one ($w_2$). We use two entailment measures: the negated KL divergence $-D_{KL}\left[\mathcal{N}(\mathbf{z}; \boldsymbol{\mu}_{w_1}, \boldsymbol{\Sigma}_{w_1}) \| \mathcal{N}(\mathbf{z}; \boldsymbol{\mu}_{w_2}, \boldsymbol{\Sigma}_{w_2})\right]$ and the cosine similarity between the means. We predict that $w_1$ entails $w_2$ if the corresponding score is above a certain threshold. As in Vilnis and McCallum (2015), the scores are optimistic, as the threshold is chosen on the test set. The thresholds are set individually for each method (including the baselines) and, hence, the comparison is fair.

Intuitively, KL should be a good choice: it would favor word pairs such that, not only their means are similar, but also such that the region, where the density function for $w_1$ is non-negligible, lies within the area where the density of $w_2$ is also high enough. Roughly speaking, level sets for $w_1$ should lie within level sets for $w_2$ (as with ('pear', 'fruit') in Figure 1).

First, we consider the entailment benchmark of Baroni et al. (2012), we will refer to it as BBDS (Table 3, left part). The results are generally consistent with the ones reported by Vilnis and McCallum. The best W2G(S) model outperforms SG. They also have the edge over BSG. We also observe that, unlike W2G, covariance information appears not to be particularly beneficial for BSG (i.e. cosine performs essentially as well as KL).

Second, we turn to the BLESS dataset (Baroni and Lenci, 2011).[8] All the considered methods perform badly on BLESS. This is even more evident from Figure 5, where we plot the histogram of the KL divergence for the entailing and not entailing pairs. Clearly, the two classes cannot be separated based on KL alone, neither for BSG, nor W2G.

We hypothesize that the reason for these, seemingly inconsistent results, is that all considered methods struggle with distinguishing hypernymy ('dog' and 'pet') from co-hyponymy ('dog' and 'cat'), as well as from other types of semantic relatedness. Making such distinction is a challenging for an unsupervised method (Weeds et al., 2014). Indeed, the BLESS dataset is harder than BBDS: it is unbalanced and contains about ten times more negatives examples than positive ones, and the negative examples are often co-hyponyms.

---
[7]The popular Gensim SG implementation is even slightly stronger than ours (0.6365), though not really comparable because of major differences in optimization.
[8]The version we are using is from Levy et al. (2015)

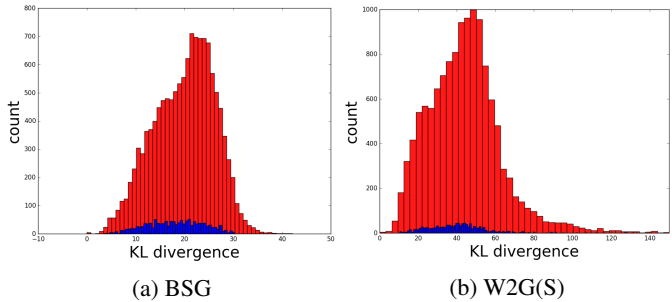

| word 1 | word 2 | KL | cosine sim. |
|---|---|---|---|
| dog | cat | 15.47 | 0.71 |
| dog | pet | 18.52 | 0.70 |
| dog | hound | 21.20 | 0.64 |
| dog | animal | 27.69 | 0.52 |

Figure 5: Histograms of KL scores on BLESS. Blue and red bars correspond to 'entailment' and 'no entailment' pairs, respectively.

Table 5: KL and cosine similarity of selected words. We used $D_{KL}\left[p(w_1)||p(w_2)\right]$, where $p(w)$ is the prior density for word $w$.

So, why does KL not work as well as we expected? Intuitively, KL can be regarded as balancing two trade-offs: penalizing for divergences between means and also penalizing for the 'wrong' type of 'inclusion' between the distributions. It is very easy to see this for the one-dimensional case, where KL can be written as: $\log \frac{\sigma_2}{\sigma_1} + \frac{\sigma_1^2 + (\mu_1 - \mu_2)^2}{2\sigma_2^2} - \frac{1}{2}$, but holds for the multivariate case as well. If the variances are sufficiently similar, then the distance between means is all that matters. We hypothesize that for the Gaussian priors within BSG and for both versions of W2G, covariances end up playing a relatively minor role. Indeed, when we checked for correlations between KL and cosine similarity, we observed that it is very strong, confirming our hypothesis. For example, dog entails animal, but dog does not entail cat, while scores in Table 5 indicate the opposite (for more examples see supplementary materials). Generally, the Pearson correlation coefficient between cosine similarity and KL equals -0.652 and -0.703, for BLESS and BBDS, respectively, indicated a very strong (linear) relation between the two. Here, we considered W2G(S), as it seemed the most promising on the basis of BBDS results, but the trend is the same for BSG and W2G(D).

Though we see that KL on its own does not seem to be sufficient for detecting entailment, we still hypothesize that it does capture information directly relevant to this task, namely, represents a degree of 'generality' of a term. In order to see that this is the case, in next subsection, we consider a modification of the entailment task. It will also hint at why W2G performs better than BSG on BBDS, while, as we will see, BSG appears to capture generality more accurately.

### 3.4 Entailment directionality prediction

In these experiments, given a pair of words, the system needs to predict if the first word entails the second one or the other way around. In other words, it is known that entailment holds for the pair but its directionality needs to be predicted. As we would like to get extra insights into results obtained in the preceding section, we extract these pairs from the same datasets, BBDS and BLESS. As the symmetric cosine similarity would be useless here, we use only KL. As a baseline, we consider the following heuristic: we compare frequencies of the two words in our corpus and predict that the less frequent word (e.g., 'kiwi') entails the more frequent one (e.g., 'fruit').

Results are presented in Table 4. First, we observe that the situations are very different for the two datasets. On BBDS, the scores are much higher for all the approaches. However, depressingly, the frequency baseline appears the strongest: neither model manages to beat it. In contrast, on BLESS, the frequency baseline is weak (only 5% higher than chance), and both W2G versions achieve results very similar to the baseline. However, BSG outperforms them by a substantial margin (10%).

These results hint at the possibility that the main reason for slightly stronger results of W2Gs in the original set-up on BBDS (Table 3) is that covariances in W2Gs capture information about token frequencies. We verify this by plotting log-determinants of covariance matrices (representing the spread of the 'ellipsoids'), as a function of token frequencies (Figure 6). The plots confirm our hypothesis: covariances for W2Gs are growing with the frequency, and hence KL with W2Gs will prefer to label frequent words as hypernyms. This is not the case at all for BSG. Note that BSG achieves similar results in directionality detection on BBDS without directly capturing frequency information (and hence 'mimicking' the frequency baseline). This, together with strong results on directionality detection for BLESS, suggests that covariances of BSG are better at capturing genuine information about the generality of a term.

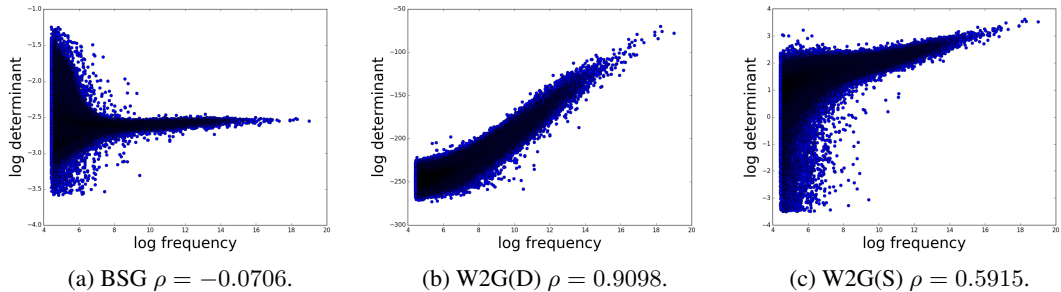

(a) BSG $\rho = -0.0706$.  (b) W2G(D) $\rho = 0.9098$.  (c) W2G(S) $\rho = 0.5915$.

Figure 6: Log determinant vs. log frequency; $\rho$ is the Pearson correlation coefficient.

### 3.5 Lexical substitution

We argued that the posterior densities in our model encode semantic properties of a word given its context. In other words, the posteriors can be used to disambiguate the word. In order to show that this is indeed the case, we consider the *lexical substitution* task, where the goal is to choose a suitable replacement for a word given its context. For example, the word 'bright' in an expression 'bright child' can be substituted with 'smart' or 'gifted' rather than 'shining'.

We used the SemEval-2007 task 10 dataset (McCarthy and Navigli, 2007). In this task, a system has to select a target word from a list of replacement candidates. We followed the set-up of Melamud et al. (2015) but kept the models the same as in the preceding sections. Namely, we did not use syntactic dependency information (shown highly beneficial in (Melamud et al., 2015)) but rather relied on a bag-of-word representation of a window of 5 words on each side. We also used 100 dimensional embeddings instead of 600 in Melamud et al. (2015).

Our model relies bag-of-word context, and the task is closely related to language modeling where representing ngrams is known to be crucial. Thus, we do not expect BSG to reach performance of state-of-the-art techniques relying on richer representations of context, such as LSTMs (e.g. (Melamud et al., 2016; Yuan et al., 2016)). We use the lexical substitution task as a way to test that the approximate posterior indeed encodes meaningful information, useful for predicting substitutions. To make the comparison fair, we use baselines which also rely on bag-of-word representations of context windows, and leave the investigation of BSG models with richer representations of context for future work.

For BSG, we ranked candidates $s$ from the set of candidates $S$ by the KL divergence: $\mathbb{D}_{KL}\left[q_\phi(\mathbf{z}|\mathbf{c},w)||p_\theta(\mathbf{z}|s)\right]$, where $q_\phi(\mathbf{z}|\mathbf{c},w)$ is the approximate posterior computed by the encoder, and $p_\theta(\mathbf{z}|s)$ is the prior distribution for the candidate word. As baselines, we used the two best performing approaches from Melamud et al. (2015), namely *Add* and *Mult*. Those heuristics score potential substitutes with respect to contexts and center words using cosine similarity of word embeddings, and mathematical operations as their names suggest. We used these heuristics on top of embeddings produced by SG and mean vectors for W2Gs and BSG. Note that they do not use covariance information.

The results (*generalized average precision*, GAP) are shown in Table 2. The encoder indeed appears substantially more effective in predicting word substitutes than the alternative methods. This shows that the posterior densities are effective at disambiguating words. In Table 6, we show top 3 substitutes proposed by the BSG encoder in several example sentences.

## 4 Related Work

Our model bears some relation to topic models: for example, one can interpret the latent meaning vector $z$ of the central word as encoding topics of the window. Topic models relying on word embeddings has been considered in the past (Das et al., 2015; Li et al., 2016). However, their modeling approaches, inference methods and applications have been quite different. Methods for inducing context-sensitive representations relying on syntactic dependency tree (using forward-backward-style algorithms) have also been studied in the past (Grave et al., 2014). Their approach is unlikely to be scalable. Very recent work of Peters et al. (2018) showed how LSTMs can be trained to provide dynamic word embeddings. These embeddings are context-aware but do not explicitly represent uncertainty about word meanings.

| Excerpts | Top 3 Substitutes |
|---|---|
| at that size it would have a **mass** of about the same as an average galaxy | conglomeration, magnitude, bulk |
| few people parallels the growing poverty of the **masses** | multitude, proletariat, throng |
| he was **really** he always wanted to be a politician | genuinely, very, definitely |
| it is his passion and to be denied that is **really** hard | absolutely, very, truly |
| between shutdown and power up when the latchup **cross** section is evaluated | sequence, segment, transverse |
| and hurting like the torments of hell for being so close to the **cross** | crucifix, sequence, angry |
| something more sedate there are quieter sophisticated **bars** in the hotels | pub, saloon, lounge |
| put granola **bars** in bowl | snack, pub, biscuit |
| i never expected to be **touched** by a weird global media event personally | affect, feel, stir |
| these stripes are continuous and do not **touch** each other | contact, finger, stroke |

Table 6: Substitutes proposed by the BSG encoder on SemEval-2007.

Reisinger and Mooney (2010) introduced a method that represents words as collections of prototype vectors. Their multi-prototype approach uses word sense discovery to partition contexts of a word and construct "sense-specific" prototypes for each cluster. Huang et al. (2012) extended this approach by incorporating global document context to learn multiple dense, low-dimensional embeddings using neural networks. This method performs word sense discrimination as a preprocessing step by clustering contexts for each word type. To model polysemous words Neelakantan et al. (2014) proposed an extension to the skip-gram model that efficiently learns multiple embeddings per word type. Their method performs word sense discrimination and embedding learning by using a nonparametric estimate of the number of senses per word type. In contrast, our model does not assume that there is a finite set of discrete senses per word. Liu et al. (2015) introduced an extension to the skip-gram model that can model the interaction between words and topics under different contexts using a tensor layer.

Our method and W2G are specifically designed to induce information about the generality of a word and capture entailment. Previous work has shown that entailment decision can also be made by 'post-processing' representations produced by standard embedding methods (e.g., (Weeds et al., 2014; Henderson and Popa, 2016)). Henderson and Popa (2016) is perhaps the most related one, as they use a probabilistic motivation for their preprocessing technique. All these post-processing methods deal with fixed rather than context-specific embeddings.

## 5 Conclusion

We introduced a method for embeddings words as probability densities. Our method produces two types of embeddings: (1) 'prior' / static embeddings representing a word type and collapsing all word senses; (2) 'posterior' / dynamic embeddings encoding a representation of a word given its context. The Gaussian embeddings of Vilnis and McCallum (2015) have been shown effective in a range of applications (e.g., modeling knowledge graphs (He et al., 2015) and cross-modal transfer from language to vision (Mukherjee and Hospedales, 2016)), our framework, providing more flexible context-sensitive alternatives, is likely to be beneficial in these settings.

## Acknowledgments

This project is supported by SAP Innovation Center Network, NWO Vidi Grant (639.022.518) and ERC Starting Grant BroadSem (678254). We would like to thank anonymous reviewers for their helpful suggestions and comments as well as Luke Vilnis for answering questions about their model.

# Supplementary Material for COLING 2018 Paper Embedding Words as Distributions with a Bayesian Skip-gram Model

**Arthur Bražinskas**[1]     **Serhii Havrylov**[2]     **Ivan Titov**[1,2]

[1]ILLC, University of Amsterdam
[2]ILCC, School of Informatics, University of Edinburgh
arthur.brazinskas@gmail.com  s.havrylov@ed.ac.uk  ititov@inf.ed.ac.uk

## 1  KL divergence and cosine similarity of semantically related words

| word 1 | word 2 | KL | cosine sim. |
|---|---|---|---|
| dog | cat | 15.47 | 0.71 |
| dog | pet | 18.52 | 0.70 |
| dog | hound | 21.20 | 0.64 |
| dog | animal | 27.69 | 0.52 |
| cappuccino | espresso | 12.59 | 0.76 |
| cappuccino | latte | 13.39 | 0.7 |
| cappuccino | coffee | 22.54 | 0.69 |
| cappuccino | drink | 30.81 | 0.54 |
| microsoft | windows | 24.41 | 0.65 |
| microsoft | google | 24.44 | 0.60 |
| microsoft | corporation | 39.40 | 0.29 |
| microsoft | company | 46.05 | 0.19 |

Table 1: KL and cosine similarity of semantically related words. We used $D_{KL}\left[p(w_1)||p(w_2)\right]$, where $p(w)$ is the prior density for word $w$.

## 2  Experimental settings

We train all our models on a concatenation of ukWaC and WaCkypedia (Baroni et al., 2009) corpora, resulting in approximately 3 billion tokens. We restricted the vocabulary size to 280,000 most frequent word types and used the sub-sampling procedure introduced in Mikolov et al. (2013) with $t = 10^{-4}$. The window size was set to 5 words on each side, the dimensionality of the embeddings and of the hidden layer of the encoder were both set to 100. The number of negative samples was equal to the number of positive ones (i.e. 10). For BSG, we used spherical covariances both for the posterior and prior densities, as they resulted in better performance in preliminary experiments: this is consistent with results reported for W2G in Vilnis and McCallum (2015). As an optimizer we used Adam (Kingma and Ba, 2014). We used large batches of 22,000 prediction tasks each. We used grid-search by running each model for 5 times on the concatenation of ukWaC and WaCkypedia dataset with different learning rates and selected the best ones based on their benchmark performance.

In order to enable fair comparison, we used our own implementation of W2G and SG, which shared the above hyperparameters with BSG.[1] The implementations of our model and the baselines will be publicly released should the paper get accepted. The initial learning rates were tuned for all models individually and set to 0.00055, 0.0065, 0.0015 and 0.0015 for BSG, W2G with the spherical covariances (WG(S)), W2G with the diagonal covariance matrix (WG(D)) and SG, respectively. All results presented in this work were obtained from one model of each type (i.e. no tuning was performed for individual benchmarks).

---

[1]The original implementation of W2G is not publicly available. Our re-implementation yields stronger results on similarity benchmarks but weaker on the entailment dataset of Baroni et al. (2012). Our implementation is also stronger across the board (including entailment) than the third party implementation used in Vulić et al. (2017) (https://github.com/seomoz/word2gauss). Athiwaratkun and Wilson (2017) also report W2G numbers very similar to ours.